\documentclass{article} 
\usepackage[final]{colm2026_conference}
\usepackage[utf8]{inputenc}
\usepackage[T1]{fontenc}
\usepackage{microtype}
\usepackage{hyperref}
\usepackage{url}
\usepackage{booktabs}
\usepackage{amsfonts}       
\usepackage{nicefrac}       
\usepackage{amsmath}
\usepackage{amssymb}
\usepackage{xcolor}         
\usepackage{fancyhdr}       
\usepackage{graphicx}       
\usepackage{tcolorbox}
\usepackage{comment}
\usepackage{placeins}
\usepackage{float}
\usepackage{adjustbox}
\usepackage{natbib}

\usepackage{lineno}

\definecolor{darkblue}{rgb}{0, 0, 0.5}
\hypersetup{colorlinks=true, citecolor=darkblue, linkcolor=darkblue, urlcolor=darkblue}

\title{Large language models reorganize representational geometry during in-context learning}

\author{Hua-Dong Xiong$^{1}$, 
Li Ji-An$^{2}$,
Robert C. Wilson$^{1,3}$,
Kwonjoon Lee$^{4}$\thanks{Co-senior author},
Xue-Xin Wei$^{5}$\footnotemark[1]
\\[1.5em]
$^{1}$ School of Psychological and Brain Sciences, Georgia Tech \\
$^{2}$ Department of Psychology, New York University \\
$^{3}$ Center of Excellence for Computational Cognition, Georgia Tech\\
$^{4}$ Honda Research Institute USA \\
$^{5}$ Departments of Neuroscience and Psychology, The University of Texas at Austin
}

\begin{document}

\ifcolmsubmission
\linenumbers
\fi

\maketitle

\begin{abstract}

Large language models (LLMs) show remarkable flexibility in adapting to novel tasks without parameter updates, a capacity known as in-context learning (ICL). Prior work has sought to understand ICL by studying the circuits, algorithms, and representations that support it. Yet why some ICL tasks are easy to solve while others are difficult remains unresolved. In this paper, we ask whether LLMs can adapt their representations arbitrarily to solve a simple linear classification task. Specifically, we construct a family of binary classification tasks in which labels are defined by projecting LLMs' own representations onto different axes. Surprisingly, although all tasks are linearly separable by construction, their in-context learnability varies systematically across axes. We find that successful ICL is accompanied by a geometric reorganization of internal representations that increases task-relevant separability. Causal interventions that amplify neural activity along the axis defining the task are insufficient to improve behavioral performance or induce this representational reorganization. We also show that LLM behavior is best described by a prototype-like algorithm operating on representations that are themselves reorganized in context to adapt to the task. Together, these findings offer a geometric account of ICL in LLMs, showing that representations acquired through training constrain what can be exploited through in-context learning.

\end{abstract}

\section{Introduction}

Large language models (LLMs) can rapidly adapt to novel tasks from a few demonstrations without parameter updates---a capability known as in-context learning (ICL) \citep{brown_language_2020}. As ICL has become a central prompting technique for steering LLM behavior, its effectiveness can vary substantially across prompts and is sensitive to the choice, ordering, and formatting of demonstrations \citep{liu_understanding_2024, wang_can_2024, zhao_calibrate_2021}.

Prior work has sought to characterize the mechanisms underlying ICL by linking it to algorithms such as online gradient descent \citep{ahn_transformers_2023, cheng_transformers_2024, vonoswald_transformers_2023, vonoswald_uncovering_2023}, Bayesian inference \citep{muller_transformers_2021, reuter_can_2025, xie_explanation_2022}, and other statistical learning methods \citep{bai_transformers_2023, chen_transformers_2022, dherin_learning_2025, garg_what_2022}. However, much of this evidence comes from small transformers trained on toy tasks, revealing what transformer architectures can learn to implement under controlled training rather than how pretrained LLMs actually adapt in context. Complementary work on pretrained LLMs has identified circuit motifs important for ICL, including induction heads \citep{olsson_incontext_2022, wang_interpretability_2022} and function vectors \citep{hendel_incontext_2023, todd_function_2024}, while representational analyses have shown that internal representations adapt to task structure during ICL \citep{park_iclr_2025, yang_provable_2025, kirsanov_geometry_2025, lampinen_linear_2026, yona_incontext_2025}. Yet identifying circuits, algorithms, and representational dynamics does not by itself explain why LLMs learn some tasks better in context than others \citep{wei_larger_2023}.

A central challenge in studying ICL is to disentangle what is learned from the demonstrations and what is learned from pretraining, such as pre-existing semantic associations and task structure \citep{min_rethinking_2022, wei_larger_2023}. This raises a more basic question: does successful ICL require reusing structure already present in the model, or can LLMs learn genuinely novel task structure from context?

To address this, we ask whether LLMs can adapt their representations arbitrarily to solve different tasks in context. We construct a family of simple binary classification tasks directly from an LLM's internal representations, assigning labels according to projections onto experimenter-selected directions. Each direction partitions the same inputs into two balanced classes. Varying the labeling direction therefore generates different target mappings over the same inputs, while systematically changing their alignment with the model's learned representational structure. By making this relationship between task structure and model representations an experimental variable, we can directly test how the model's existing representational structure constrains ICL. This setup also connects to the neuroscience view of classification as representational untangling \citep{dicarlo_untangling_2007, dicarlo_how_2012, yamins_using_2016} (Fig.~\ref{fig:diagram}), allowing us to test whether LLMs can similarly untangle task-relevant representations along arbitrary directions during ICL.

We vary the labeling axis that defines each ICL task along principal component directions of decreasing variance and test how the choice of labeling axis shapes both ICL behavior and online representational change. We find that experimenter-defined label partitions aligned with high-variance representational directions are learned more rapidly and reach higher levels of performance than those defined along low-variance directions. Thus, ICL does not operate equally well over arbitrary task mappings: what can be learned in context is constrained by the model's existing representational structure.

To characterize the learning dynamics underlying these differences, we use manifold capacity theory \citep{chung_classification_2018, cohen_separability_2020} to quantify how the geometry of the experimenter-defined class manifolds evolves as in-context examples accumulate. We find that successful ICL is accompanied by an online geometric reorganization of final-layer representations: manifold capacity increases, radius and dimension decrease, and the induced classes become progressively more separable. A causal intervention that amplifies variation along the labeling axis does not improve either ICL behavior or representational reorganization, indicating that task-aligned activity gain alone is insufficient for successful ICL. Finally, we compare the model's behavior with a range of online learning algorithms operating on the same final-layer representations. We find that LLM responses are best matched by a prototype-like algorithm, suggesting that, in this setting, models integrate contextual evidence in a manner consistent with class-level summaries.

Together, these findings provide a geometric and computational account of ICL in pretrained LLMs, showing that ICL is both constrained by and reflected in the geometry of internal representations. More broadly, our representation-defined paradigm provides a controlled way to study which task mappings pretrained LLMs can learn in context and how their representations change as those mappings are acquired.

\section{Methods} \label{sec:methods}

\begin{figure}[!htb]
\centering
\includegraphics[width=1\textwidth]{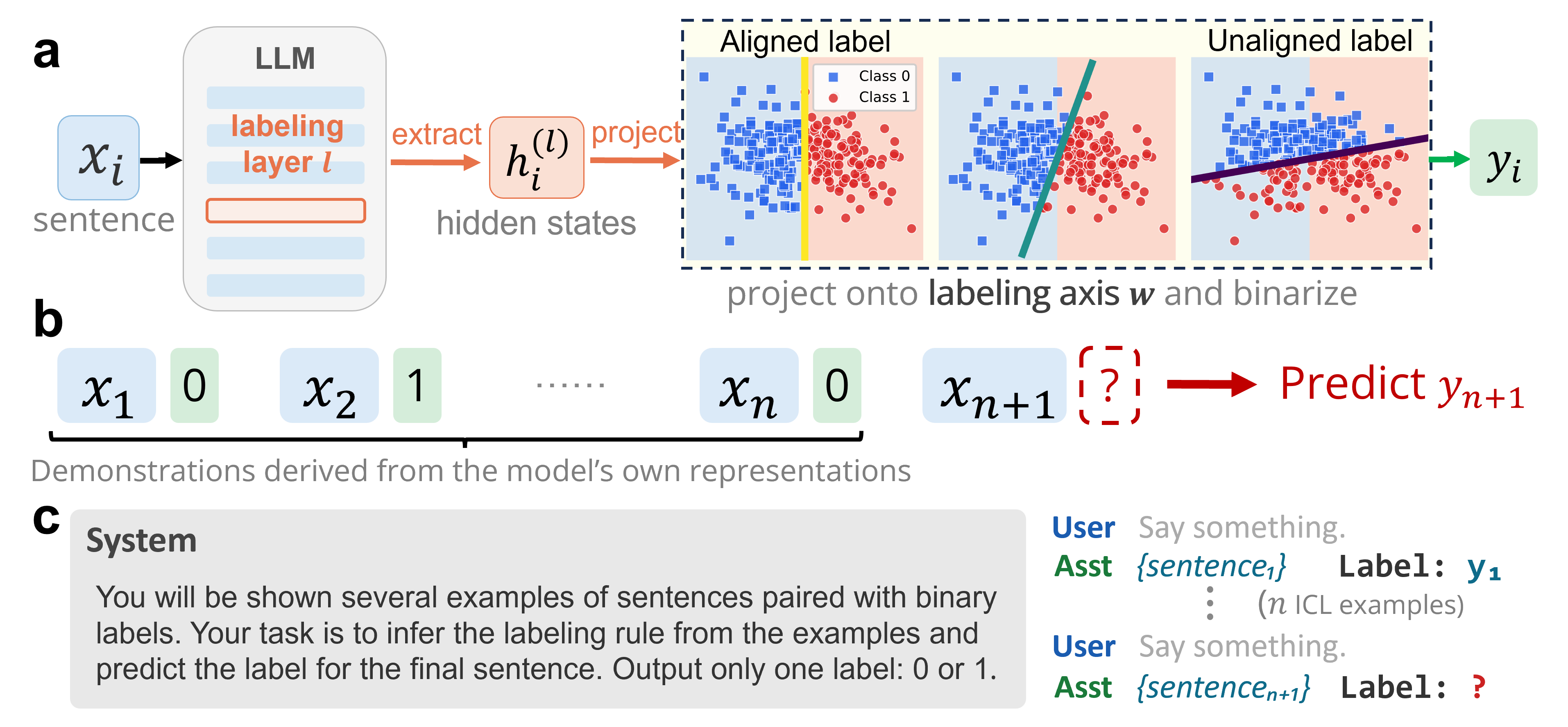}
\vspace{-15pt}
\caption{\textbf{Representation-defined in-context classification task.}
\textbf{(a)} For each sentence $x_i$, we extract a sentence-level representation $h_i^{(\ell)}$ from a selected labeling layer $\ell$ and assign the label $y_i = \operatorname{sign}(w^\top h_i^{(\ell)} + b)$, where $w$ defines the labeling axis and $b$ is chosen to approximately balance the two classes. Background shading shows the decision regions of a logistic-regression (LR) classifier trained on the same representations to predict the dataset labels; we refer to its weight vector as the LR axis. Point colors indicate the labels assigned by $w$. As $w$ becomes increasingly misaligned with the LR axis, the induced labels increasingly differ from the dataset labels. Experimental labeling conditions are constructed at five layers sampled uniformly across model depth.
\textbf{(b)} The model is given a sequence of sentence--label pairs $(x_i,y_i)$ and asked to predict the label of a new sentence.
\textbf{(c)} Example task prompt.}
\label{fig:neurofeedback}
\end{figure}

\subsection{Representation-defined in-context classification task}
\label{sec:nf_task}

We construct a family of in-context classification tasks whose labels are defined directly from a pretrained LLM's hidden representations~\citep{sadtler_neural_2014, ji-an_language_2025} (Fig.~\ref{fig:neurofeedback}). Let \(\mathcal{D}=\{x_i\}_{i=1}^N\) denote a dataset of sentences and \(M\) a trained LLM. For each labeling layer \(\ell\), we extract a sentence-level representation \(h_i^{(\ell)}\in\mathbb{R}^d\) by mean-pooling residual-stream activations over the assistant-response span, excluding the assistant header and the final end-of-sequence token (results are similar when using the final response token). Throughout, we use $h$ for hidden representations and specify the relevant layer, position, and context locally.

From these representations, we construct a set of candidate labeling axes
\(\mathcal{W}(\mathcal{D},M,\ell)\subset\mathbb{R}^d\). Each axis
\(w\in\mathcal{W}(\mathcal{D},M,\ell)\), together with an offset \(b_{\ell,w}\), defines the affine score and binary partition
\[
a_i^{(\ell,w)}=\langle w,h_i^{(\ell)}\rangle+b_{\ell,w},
\qquad
y_i^{(\ell,w)}=\mathbf{1}\!\left[a_i^{(\ell,w)}>0\right].
\]
Here, the labeling layer \(\ell\) determines the representation space in which the task is defined, while the labeling axis \(w\) determines the partition within that space. Each pair \((\ell,w)\) therefore defines a distinct classification task with its corresponding label assignment \(y^{(\ell,w)}\).

At evaluation time, the model receives \(n\) labeled in-context examples,
\((x_1,y_1^{(\ell,w)}),\ldots,(x_n,y_n^{(\ell,w)})\), and is asked to predict
\(y_{n+1}^{(\ell,w)}\) for a new sentence \(x_{n+1}\) (Fig.~\ref{fig:neurofeedback}c; full prompts see Appendix~\ref{appendix:prompts}). 

We consider two types of labeling axes: a logistic-regression (LR) axis and principal-component (PC) axes. We obtain the LR axis by fitting a linear classifier to predict the dataset's ground-truth semantic labels from the representations \(h_i^{(\ell)}\), and refer to its weight vector as the LR axis. This provides a reference direction aligned with task-relevant semantic structure in the model's representations. We compute the PC axes from the same representations and define labels by thresholding the centered projection onto each PC at zero. We consider \(k\in{1,2,4,8,32,128,512}\).

PC axes provide an ordered family of labeling directions, ranked by explained variance in the LLM's representation space. Although all PC-defined tasks are linearly separable by construction, they differ in where the target partition lies along this representational spectrum. We therefore use PC rank to probe how ICL performance varies across representational directions.

We define these tasks at five labeling layers spanning model depth, chosen at evenly spaced layer quantiles \((0,0.25,0.5,0.75,1)\) (details see Appendix~\ref{sec:nf_axes}). Combining eight labeling axes (one LR axis and seven PC axes) with five labeling layers produces 40 distinct representation-defined task conditions per LLM. We evaluate each task condition on 100 sampled sequences under two counterbalanced label-token assignments, swapping which class is represented by ``0'' versus ``1'' to control for biases associated with the output tokens, yielding 200 trials per task condition in total (Appendix~\ref{sec:nf_trials}). Unless otherwise noted, results are averaged across the five labeling layers; layer-specific results are similar (see Appendix).

\subsection{Models and datasets}

We consider two evaluation settings. (1) We assess model-level generalization by aggregating results across eight instruction-tuned models: Llama3 \citep{grattafiori_llama_2024} (3B, 8B), Gemma3 \citep{gemma_gemma_2025} (1B, 4B, 12B, 27B), and Qwen3 4B \citep{yang_qwen3_2025} and Qwen3.5 27B \citep{qwen_qwen35_2026}. All models are evaluated on ETHICS-commonsense \citep{hendrycks_aligning_2020}. (2) To evaluate task-level generalization, we fix a single model, Qwen3 4B, and evaluate it across five sentence classification datasets: ETHICS-commonsense and ETHICS-justice \citep{hendrycks_aligning_2020}, SST-2 and CoLA from GLUE \citep{wang_glue_2018}, and LIAR \citep{wang_liar_2017} (details in Appendix~\ref{sec:dataset}).

\section{Results}

\subsection{ICL performance varies systematically across labeling axes}

\begin{figure}[!htb]
\centering
\includegraphics[width=1.0\textwidth]{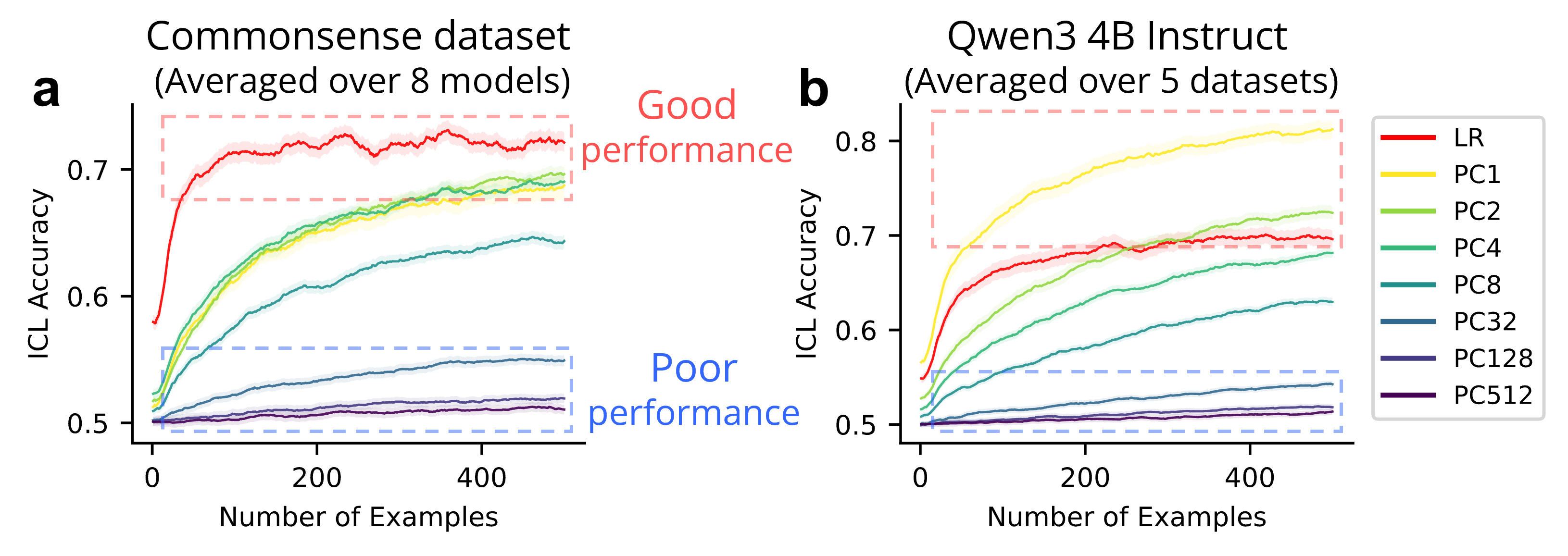}
\vspace{-15pt}
\caption{
\textbf{ICL performance varies systematically across labeling axes that define the task.}
ICL accuracy as a function of the number of in-context examples. Tasks defined by the LR axis and leading PCs are learned more rapidly and reach higher performance, whereas tasks defined by trailing PCs exhibit slower learning and lower saturation levels.
\textbf{(a)} Accuracy averaged over five labeling layers sampled at evenly spaced depth quantiles and across eight models of varying sizes and families on ETHICS-commonsense. Shaded regions denote $\pm 1$ standard error of the mean (SEM) across models.
\textbf{(b)} Accuracy for Qwen3 4B averaged over the same five labeling layers and across five datasets. Shaded regions denote $\pm 1$ SEM across datasets.
}
\label{fig:nficl_perf}
\end{figure}

We study ICL performance in two settings: ETHICS-commonsense, averaged across eight pretrained LLMs (Fig.~\ref{fig:nficl_perf}a), and Qwen3 4B, averaged across five datasets (Fig.~\ref{fig:nficl_perf}b). We find that ICL performance improves as more demonstrations are provided (Fig.~\ref{fig:nficl_perf}), indicating that LLMs can use contextual examples to infer the experimenter-defined label mapping in context.

However, ICL performance varies systematically with the labeling axis. The same axis ordering appears across all five labeling layers (Appendix~\ref{sec:acc_by_layer}, Figs.~\ref{fig:acc_by_layer_commonsense}--\ref{fig:acc_by_layer_qwen3_4b}). As the number of in-context examples increases, tasks defined by the LR axis and leading principal components (PC1, PC2, PC4, and PC8) improve rapidly and approach higher performance plateaus, whereas tasks defined by trailing components (PC32, PC128, and PC512) improve more slowly and approach lower plateaus over the evaluated context lengths. Thus, even though all tasks are generated by linear partitions of the model's own hidden representations and are linearly separable by construction, they differ markedly in their learning dynamics.

Previous work on finite-sample spectral learning theory \citep{canatar_spectral_2023, yang_provable_2025} provides a partial account of this pattern. For a PC-defined partition, the label--representation signal along the labeling direction is proportional to the variance of that PC coordinate. Trailing-PC partitions therefore yield weaker signals and are harder to identify from finite demonstrations (see Appendix~\ref{sec:finite_sample_pc_spectrum} for details). However, this account does not explain the differences in observed ceiling performance across axes or why learning rates do not strictly follow PC variance.

\subsection{Manifold geometry quantifies representational untangling}

Having shown a graded in-context learnability landscape at the behavioral level, we next ask what neural mechanisms underlie these differences in ICL performance. Inspired by the view of classification as neural manifold untangling \citep{dicarlo_untangling_2007, dicarlo_how_2012, yamins_using_2016} (Fig.~\ref{fig:diagram}), we test whether successful ICL makes the two induced class manifolds increasingly separable.

\begin{figure}[!htb]
\centering
\includegraphics[width=0.8\textwidth]{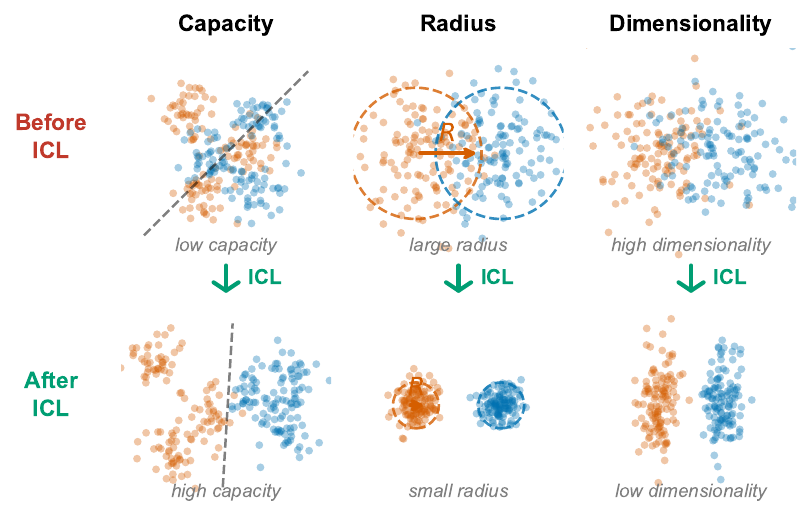}
\vspace{-15pt}
\caption{\textbf{Schematic of geometric signatures of neural manifold untangling during learning.} Each column illustrates one geometric property of class manifolds before (top) and after (bottom) successful untangling. \emph{Capacity}: overlapping manifolds have low capacity and are difficult to linearly separate; separating the manifolds increases capacity. \emph{Radius}: large within-class spread hinders separation; compressing the manifolds reduces radius. \emph{Dimension}: high-dimensional within-class variability complicates separation; contraction onto fewer effective directions reduces dimension.}
\label{fig:geometry}
\end{figure}

We quantify how final-layer sentence representations, from which the model's output is read out, reorganize during ICL using geometric measures of these class manifolds. Let $h_{r,f,t}$ denote the context-conditioned final-layer representation at position $t$ of sampled sequence $r$ under label-token assignment $f$. At each position, the labels split these representations into two point clouds, $\mathcal{M}_{c,t}=\{h_{r,f,t}\mid y_{r,f,t}=c\}$ for $c\in\{0,1\}$; all metrics below describe the geometry of these two clouds, rather than the labeling axis itself. Specifically, we use metrics from manifold capacity theory \citep{chung_classification_2018, cohen_separability_2020} and the signal-to-noise ratio (SNR) between manifolds \citep{sorscher_neural_2022}. Manifold capacity quantifies how readily class manifolds can be linearly separated: higher capacity indicates a geometry that better supports linear classification. Radius and dimension capture complementary aspects of within-class variability---broader and higher-dimensional manifolds are generally more difficult to separate. SNR quantifies how clearly discriminative structure emerges from background variability (Fig.~\ref{fig:geometry}). For each of the five labeling layers $\ell$ and each labeling axis (LR or $\mathrm{PC}_k$), we track these metrics for the induced class manifolds in the model's final-layer representations across successive in-context learning steps ($t=1,2,\ldots$), using 200 runs at each step. Because the qualitative patterns are consistent across layers, we report averages over the five labeling layers here (details are provided in Appendix~\ref{sec:appendix_geometry}).

\subsection{Online reorganization of representational geometry during ICL}

Why does ICL performance differ across labeling axes, succeeding for the LR axis and leading PCs but performing poorly for trailing ones? Under the manifold-untangling framework, successful ICL should be accompanied by increasing linear separability (higher capacity), decreasing within-class complexity (lower radius and/or dimension), and improving discriminability (higher SNR) as demonstrations accumulate.

\begin{figure}[!htb]
\centering
\includegraphics[width=1\textwidth]{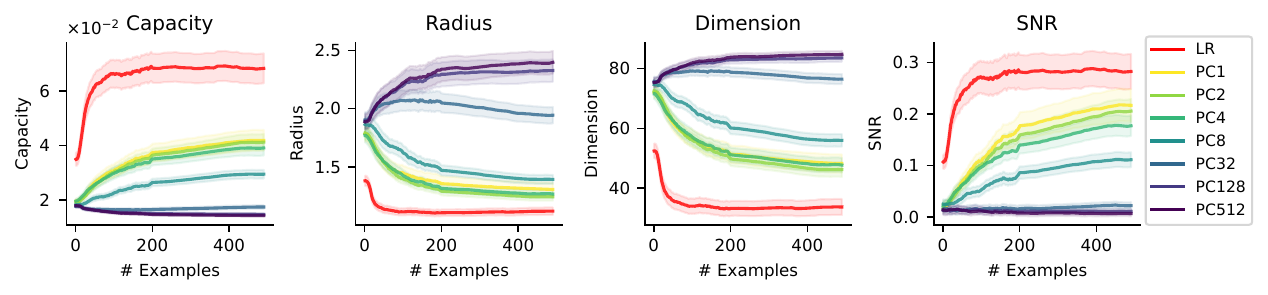}
\vspace{-15pt}
\caption{\textbf{Geometric reorganization of neural representations during ICL.}
Capacity, radius, dimension, and SNR of the final-layer representations across in-context examples, averaged over five labeling layers and eight models. Each curve denotes a different labeling axis.
The LR axis and leading PCs (PC1/2/4/8) exhibit strong geometric reorganization with in-context examples, with increasing capacity and SNR and decreasing radius and dimension, whereas trailing PCs (PC32/128/512) remain in low-capacity, low-SNR regimes, showing little geometric reorganization and even increased radius and dimension.
}
\label{fig:icl_geometry_axis}
\end{figure}

We test these predictions by systematically analyzing manifold capacity, radius, dimension, and SNR as a function of the number of in-context examples (Fig.~\ref{fig:icl_geometry_axis}). We compute each metric from final-layer representations and average over the five labeling layers and experimental runs for each labeling axis.

For labeling axes with strong ICL performance (the LR axis and leading PCs), capacity increases while radius and dimension decrease with the number of examples, and SNR rises sharply. Rising capacity and SNR together with falling radius and dimension indicate progressive untangling of the two class manifolds in the final-layer representations. For more difficult labeling axes (trailing PCs), the geometric changes are substantially attenuated: capacity remains low, radius and dimension change little or even increase slightly, and SNR remains near zero, indicating limited movement toward a linearly separable configuration. Final-layer geometry therefore mirrors the success and failure regimes in Fig.~\ref{fig:nficl_perf}. The axis-dependent reorganization holds across labeling layers and models and also appears for Qwen3 4B across five datasets (Appendix Fig.~\ref{fig:icl_geometry_qwen3_4b}).

These results link ICL success to pretrained representational structure: the model's ability to untangle the two class manifolds depends on the labeling layer and labeling axis that define what must be learned in context. Along trailing-PC axes, final-layer representations resist untangling and show little of the geometric reorganization associated with successful ICL, even after many demonstrations. Along the LR axis and leading PCs, by contrast, the model reliably reorganizes its representations toward greater class separability. ICL thus reflects a constrained form of online representational untangling whose success depends on how readily pretrained representations support the target partition.

\subsection{Representational geometry predicts ICL performance}
\label{sec:results_corr}

We next ask how representations connect to behavior: specifically, whether geometric reorganization tracks the improvement in behavioral performance during ICL. We correlate ICL accuracy with final-layer representational geometry metrics across analysis units spanning experimental conditions, models, and in-context example counts on ETHICS-commonsense (Table~\ref{tab:geometry_perf}). Each analysis unit is a specific combination of model, labeling axis, labeling layer, and number of in-context examples; within each unit, accuracy and geometry metrics are computed across repeated runs (details in Appendix~\ref{sec:appendix_geometry}). Final-layer geometry strongly predicts ICL performance: accuracy increases with manifold capacity ($r=0.83$) and SNR ($r=0.71$) and decreases with radius ($r=-0.78$) and especially dimension ($r=-0.87$). Capacity, radius, dimension, and SNR thus provide a compact geometric characterization of ICL success, quantitatively linking the extent of representational untangling to behavioral performance.

\begin{table}[htb]
\centering
\caption{\textbf{Representational geometry predicts behavioral performance during ICL.} Pearson $r$ between each geometric metric and ICL accuracy, aggregated across axes and in-context positions. $^{***}\,p < .001$.}
\label{tab:geometry_perf}
\begin{tabular}{cccc}
\toprule
Capacity & Radius & Dimension & SNR \\
\midrule
$0.83^{***}$ & $-0.78^{***}$ & $-0.87^{***}$ & $0.71^{***}$ \\
\bottomrule
\end{tabular}
\end{table}

\subsection{Amplifying task-aligned activity is insufficient to improve ICL}
\label{sec:causal_intervention}

The connection between neural representations and behavior, together with the observed spectral bias, raises a natural question: can differences in learnability be largely explained by spectral bias, such that axes with larger variance are easier to learn? We tested this causally by holding the labeling axis and induced labels fixed while amplifying centered projections along that axis. At a given labeling layer, we multiplied the centered projection onto the PC axis by $\gamma=2$, quadrupling its variance without redefining the task (we also swept selected values of $\gamma$; see Appendix~\ref{sec:causal_axis_gain} for details). If task-aligned variance were sufficient, this manipulation should make the induced class manifolds easier to untangle and improve ICL. We applied the intervention to PC2, PC8, PC32, and PC128 across the five labeling layers in three models. For accuracy and each final-layer geometry metric, we report the resulting difference, $\Delta$, between each intervention run and its matched non-intervened baseline.

Amplifying activity along the labeling axis did not improve ICL or facilitate representational reorganization. Behavioral accuracy was unchanged or slightly lower across all tested axes, and although the intervention modestly inflated radius and dimension, it did not reliably raise capacity or SNR (Table~\ref{tab:causal_axis_gain_by_axis}). The causal effect on the in-context learning gain, defined as the accuracy difference between the last and first 50 evaluated positions, also remained near zero. The same null pattern held when the deltas were broken down by labeling-layer depth (Appendix Table~\ref{tab:causal_axis_gain_by_labeling_layer}). Thus, amplifying activity along a fixed labeling axis is insufficient to improve online untangling; the difficulty ordering cannot be reduced to rescaling activity along that direction.

\begin{table}[!htb]
\centering
\small
\setlength{\tabcolsep}{6pt}
\caption{\textbf{Amplifying activity along the labeling axis does not improve ICL.} Mean causal-minus-baseline change ($\Delta$) after applying a $\gamma=2$ centered gain along the labeling PC axis, by labeling axis. $\Delta G_{\mathrm{ICL}}$ is the causal-minus-baseline change in $G_{\mathrm{ICL}}=A_{\mathrm{last\,50}}-A_{\mathrm{first\,50}}$. Results are averaged across five labeling layers and three models on ETHICS-commonsense; values are rounded to three decimals.}
\label{tab:causal_axis_gain_by_axis}
\begin{adjustbox}{max width=\textwidth}
\begin{tabular}{ccccccc}
\toprule
Axis & $\Delta$ ICL accuracy & $\Delta$ capacity & $\Delta$ radius & $\Delta$ dimension & $\Delta$ SNR & $\Delta G_{\mathrm{ICL}}$ \\
\midrule
PC2 & $-0.002$ & $-0.001$ & $0.013$ & $0.845$ & $-0.008$ & $0.006$ \\
PC8 & $-0.002$ & $0.000$ & $0.020$ & $0.668$ & $-0.004$ & $-0.001$ \\
PC32 & $-0.005$ & $0.000$ & $0.033$ & $0.951$ & $-0.002$ & $-0.007$ \\
PC128 & $-0.002$ & $0.000$ & $0.006$ & $0.147$ & $-0.002$ & $0.001$ \\
\bottomrule
\end{tabular}
\end{adjustbox}
\end{table}

\subsection{Prototype learning best describes ICL behavior over reorganized representations}

\begin{figure}[!htb]
\centering
\includegraphics[width=0.8\textwidth]{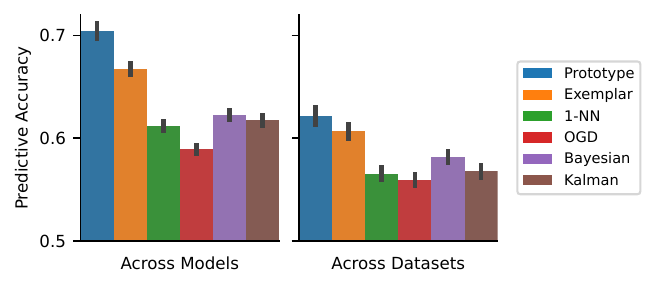}
\vspace{-8pt}
\caption{\textbf{Comparison of different online algorithms in predicting LLM behavior.}
Accuracy of six different online algorithms in predicting LLM behavior (descriptive fit), computed from final-layer representations and averaged over five labeling layers. The prototype model matches LLM behavior best.
\textbf{(a)} Model fitting accuracy averaged over eight LLMs on ETHICS-commonsense. Shaded regions denote $\pm 1$ SEM across models.
\textbf{(b)} Model fitting accuracy for Qwen3 4B averaged over five datasets. Shaded regions denote $\pm 1$ SEM across datasets. }
\label{fig:cog_models}
\end{figure}

Finally, previous studies have characterized LLM behavior by fitting algorithms purely at the behavioral level. Here, having shown that internal representations change systematically over the course of ICL, we ask what algorithm operates on these changing representations. We therefore compared the model's predictions with several online learning algorithms operating on the same final-layer representations (Fig.~\ref{fig:cog_models}). These learners differ in how they represent and update information from past examples.

Casting binary classification as accumulating evidence for two classes, these classic online models differ mainly in what they store from past demonstrations and how they update it. Prototype learning summarizes each class by a single centroid in normalized final-layer representation space and assigns a new example to whichever class centroid is closer. Exemplar learning, rather than collapsing examples into a centroid, compares a new example against all memorized instances to decide which class is closer. The nearest-neighbor model is the simplest such variant, predicting from the single closest stored instance (1-NN). Online gradient descent (OGD) instead maintains a linear decision rule and nudges its weights after each example in proportion to the current prediction error. The Bayesian class-mean model also represents each class by a mean, but treats that mean as an uncertain latent variable and updates both its estimate and uncertainty online. The Kalman filter shifts the latent state from class means to the weights of a linear decoder, updating both the decoder and its uncertainty with a gain that depends on how informative the current example is (details in Appendix~\ref{sec:fit_cog_model}).

Running each model on the same demonstrations, we measure how often its trial-by-trial predictions agree with the LLM's own responses (descriptive fit; averaged across labeling layers, details in Appendix~\ref{sec:fit_cog_model}). We find that prototype learning best matches the LLM's responses (Fig.~\ref{fig:cog_models}). The prototype fit suggests that the LLM behaves like a centroid-based learner, summarizing each class by its mean final-layer representation---a representation that is not fixed but is itself progressively reorganized to become more separable as demonstrations accumulate. Notably, more flexible linear online learners can achieve higher task accuracy from the same representations. Thus, ICL in our setting is best described by a prototype-like learner operating over representations that themselves reorganize during in-context learning.

\section{Discussion}
\label{sec:discussion}

In this work, we first asked what determines whether a task is learnable in context. We addressed this question using a family of in-context binary classification tasks whose labels are defined by linear partitions of an LLM's own hidden representations. Although every task is linearly separable by construction, in-context learnability varies systematically across partitions. Tasks defined by the LR axis and leading PCs are learned more rapidly and attain higher accuracy within the evaluated context range, whereas trailing-PC tasks improve more slowly and remain less accurate. We then asked how representations change when learning succeeds. We found that successful ICL is marked by online manifold untangling: final-layer representations become progressively more separable as in-context examples accumulate. The gain intervention further shows that amplifying variation along a labeling axis is insufficient to improve behavior or induce this geometric reorganization. Finally, a prototype-like learner best describes behavior over representations that are themselves reorganized by context.

Prior work on representational adaptation typically begins with task structure specified by the experimenter and asks how internal representations adapt to it \citep{park_iclr_2025, yang_provable_2025, kirsanov_geometry_2025, lampinen_linear_2026, yona_incontext_2025}. We invert this experimental design by using the pretrained representation space itself to define the learning problem. This allows us to test how the choice of labeling axis shapes both in-context learnability and online representational change.

\paragraph{Pretrained representations shape an in-context learnability landscape.} All tasks have the same form and are linearly separable by construction, yet they differ markedly in sample efficiency and in the accuracy attained over the evaluated context range. A simple finite-sample spectral account offers a partial explanation: leading PCs have higher variance, producing stronger label-aligned signal that can be recovered from fewer examples \citep{canatar_spectral_2023} (Appendix~\ref{sec:finite_sample_pc_spectrum}). This account can explain slower improvement for trailing-PC tasks, but it neither reproduces the learning curves exactly nor predicts differences in observed ceiling performance. The gain intervention further shows that amplifying variation along a fixed labeling axis is insufficient to improve ICL, so axis-wise magnitude alone cannot account for the observed learnability landscape. More broadly, the systematic advantage of leading PCs and the LR axis suggests that partitions better supported by representational structure inherited from training are more readily learned in context. In this sense, pretrained representational structure constrains the mappings that ICL can readily acquire.

\paragraph{A representational and algorithmic account of ICL.} Our findings complement prior work that interprets ICL in terms of online algorithms that transformers may implement during inference \citep{ahn_transformers_2023, cheng_transformers_2024, vonoswald_transformers_2023, vonoswald_uncovering_2023, muller_transformers_2021, reuter_can_2025, bai_transformers_2023, xie_explanation_2022}. Related analyses of trained LLMs have described ICL as Bayesian belief updating \citep{bigelow_belief_2026}. Our focus is complementary: we jointly characterize how representations reorganize and which tested candidate rule best predicts responses from those context-conditioned representations. At the behavioral level, LLM responses are best described by a prototype-like learner operating on final-layer representations that are themselves reorganized by context. The prototype rule is also closely connected to Hebbian learning: for balanced binary classes, the difference between the two empirical class prototypes is proportional to a label-signed sum of the representations \citep{hebb_organization_1949,mcclelland_distributed_1985} (Appendix~\ref{sec:cognitive_models}). Thus, the geometric analyses and algorithm fitting connect how representations change during ICL to how those representations are used to produce behavior. This account also extends work relating fixed representational geometry to few-shot object classification \citep{sorscher_neural_2022}: in our study, representations themselves reorganize \textit{online} as in-context examples accumulate.

\paragraph{Neural constraints on learning.} The dependence of ICL on pretrained representations parallels constraints on rapid learning in neuroscience. Monkeys learn brain--computer interface mappings within the intrinsic manifold of existing population activity more readily than mappings that require activity outside it \citep{sadtler_neural_2014}. Short-term improvement can arise by reassociating a largely fixed set of activity patterns \citep{golub_learning_2018}, whereas multi-day learning can produce new activity patterns \citep{oby_new_2019}. Without weight updates, ICL may likewise be constrained by the representational repertoire established during pretraining rather than being free to construct arbitrary task-relevant structure. This analogy offers one possible explanation for why trailing-PC tasks show limited improvement over the evaluated context lengths: partitions that are poorly supported by pretrained representations may remain difficult to learn from context alone. The analogy concerns a common constraint on rapid learning, not a shared mechanism.

\paragraph{Limitations and future directions.} Several limitations follow from our framing and point to natural next steps. Linear partitions afford strong experimental control but leave open how far the same principles extend to generation, hierarchical or structured latent variables, and nonlinear task structure. Our labeling-axis families are likewise principled but not unique: logistic regression identifies salient label-related directions and principal components define an ordered family indexed by explained variance, yet these probes do not exhaust the possible notions of task-relevant geometry, and alternatives such as random directions or nonlinear manifold coordinates may expose different constraints and offer finer experimental control. The observed ceiling performance of trailing-PC tasks over the evaluated context lengths is an empirical ceiling, not necessarily an absolute one, because performance may continue to improve with additional examples. The association between geometric measures and ICL performance also remains correlational. The gain intervention tests whether amplifying variation along a labeling axis is sufficient to improve ICL, not whether manifold untangling is causally necessary or which properties of an axis make its induced partition more learnable. Likewise, the online-learner comparison is descriptive and does not establish that the LLM literally implements a prototype algorithm. A further question is how the geometry that supports or limits ICL emerges during pretraining and is modified by instruction tuning or other finetuning. Extending the framework to reasoning settings is also promising: chain-of-thought generation may enable additional sequential computation that lets the model further reorganize task-relevant representations before committing to an answer. Finally, a formal theory linking pretrained representational structure, online manifold reorganization, and behavioral readout remains to be developed.

\paragraph{Conclusion.} Together, our results answer the two questions that motivated this study. First, pretrained representations shape which experimental conditions are readily learned from in-context examples. Second, when learning succeeds, final-layer representations become progressively more separable as context accumulates. Among the online learners tested, a prototype-like readout best describes LLM responses over these context-dependent representations. ICL can therefore be understood as rapid adaptation constrained by the representational structure inherited from pretraining.

\clearpage


\section*{Acknowledgments}
This work was supported by Lambda Research Grant awarded to H.-D.X. and Sloan Research Fellowship awarded to X.-X.W.

\section*{Reproducibility Statement}
Code for reproducing all experiments and analyses is available at \url{https://github.com/sakimarquis/neural-constraints-icl}.


\clearpage
\bibliography{references}
\bibliographystyle{colm2026_conference}

\clearpage
\appendix
\section{Appendix}

\subsection{Models}
\label{sec:models}

We use eight instruction-tuned open-weight models from three families spanning 1B to 27B parameters:
\begin{itemize}
    \item \textbf{Llama 3} \citep{grattafiori_llama_2024}: Llama-3.2-3B-Instruct and Llama-3.1-8B-Instruct.
    \item \textbf{Gemma3} \citep{gemma_gemma_2025}: Gemma-3-1B-IT, Gemma-3-4B-IT, Gemma-3-12B-IT, and Gemma-3-27B-IT.
    \item \textbf{Qwen 3} \citep{yang_qwen3_2025, qwen_qwen35_2026}: Qwen3-4B-Instruct-2507 and Qwen3.5 27B.
\end{itemize}

``B'' denotes the number of parameters in billions. We run all models with thinking (extended reasoning) disabled. For each model, we extract hidden-state representations from its residual stream (Appendix~\ref{sec:nf_axes}) and use the same model as the in-context learner that predicts the experimental labels. The cross-model analyses aggregate all eight models on ETHICS-commonsense, whereas the cross-dataset analyses fix Qwen3-4B-Instruct-2507 and vary the dataset.

\subsection{Datasets}
\label{sec:dataset}

We use two dataset configurations. The cross-model analyses evaluate all models on ETHICS-commonsense \citep{hendrycks_aligning_2020}. The cross-dataset analyses fix Qwen3-4B-Instruct-2507 and evaluate five binary sentence-classification datasets: ETHICS-commonsense, ETHICS-justice \citep{hendrycks_aligning_2020}, SST-2, CoLA \citep{wang_glue_2018}, and LIAR. For LIAR, we retain only statements labeled \texttt{true} or \texttt{false}, map them to 1 and 0, respectively, and discard the four intermediate categories. Across datasets, we retain only binary labels, construct balanced train/test partitions, and exclude sentences shorter than 5 words or longer than 30 words to keep prompt lengths comparable across experimental conditions.

\subsection{Representation-defined ICL task construction}

\subsubsection{Constructing labeling axes}
\label{sec:nf_axes}

We construct a spectrum of in-context classification tasks by varying the labeling axis across two families: LR and PCs. The LR family contains a single logistic-regression (LR) axis obtained by fitting a linear classifier to dataset labels (e.g., morality in ETHICS) using $h_i^{(\ell)}$ as features. As the optimal linear direction for recovering dataset labels from a given labeling layer's sentence representations, the LR axis provides a supervised semantic reference.

For the PC family, we compute principal components (PCs) of centered residual-stream activations over dataset examples at each layer. These axes let us test how variance structure in the labeling-layer representation space shapes ICL. Explained variance orders the PCs along a controlled spectrum: although every PC-defined task is linearly separable by construction, high-variance directions tend to be learned more readily from finite samples, whereas trailing directions are harder to infer. Most PCs align only modestly with the dataset-specific LR axis and therefore overlap little with that supervised direction. Across the PC family, increasing rank corresponds to decreasing explained variance and empirically increasing task difficulty. PC512, for example, defines a partition along a low-variance direction of the preparation-set covariance.

We now restate the task construction in the notation of Section~\ref{sec:nf_task}. Fix a labeling layer $\ell$ and a labeling axis $w \in \mathcal{W}(\mathcal{D},M,\ell)$. The affine score is
\[
a_i^{(\ell,w)}=\langle w,h_i^{(\ell)}\rangle+b_{\ell,w},
\]
and the induced label is
\[
y_i^{(\ell,w)}=\mathbf{1}[a_i^{(\ell,w)}>0].
\]
For LR-defined tasks, $b_{\ell,w}$ is the fitted intercept. For PC-defined tasks, $b_{\ell,w}=-\langle w,\bar h^{(\ell)}\rangle$, where $\bar h^{(\ell)}$ is the mean used to fit PCA, and hence $a_i^{(\ell,w)}=\langle w,h_i^{(\ell)}-\bar h^{(\ell)}\rangle$. The zero threshold therefore defines a hyperplane through the centered PCA origin. Because centering does not guarantee equal class sizes, we quantify the canonical positive-label fraction before label-token counterbalancing as $\pi_{\ell,w}=N_{\mathrm{eval}}^{-1}\sum_i\mathbf{1}[a_i^{(\ell,w)}>0]$. The two label-token assignments balance the displayed score tokens but do not change this fraction. For Qwen3-4B on LIAR, the fraction ranges from 0.457 to 0.534 across PC-defined experimental conditions. Although alternative geometric constructions (e.g., nonlinear manifold geometry) are possible, PCA provides a linear, ordered family of directions that is not explicitly optimized for the dataset labels.

\subsubsection{Finite-sample intuition for the PC spectrum}
\label{sec:finite_sample_pc_spectrum}

Linear separability guarantees that a separator exists, not that a finite-sample learner can identify it equally quickly along every direction. This spectral bias favors high-eigenvalue modes, which are recovered from fewer samples \citep{canatar_spectral_2023}. To express the intuition in our notation, fix a labeling layer $\ell$ and define the centered representation $\delta h_i^{(\ell)}=h_i^{(\ell)}-\bar h^{(\ell)}$. Under an idealized Gaussian approximation, let $\delta h_1^{(\ell)},\ldots,\delta h_n^{(\ell)}\overset{\mathrm{i.i.d.}}{\sim}\mathcal{N}(0,\Sigma_\ell)$, let $\{\lambda_{\ell,j}\}_{j=1}^d$ denote the full covariance spectrum, and let $w=\mathrm{PC}_k$ be a unit eigenvector satisfying $\Sigma_\ell w=\lambda_{\ell,k}w$. The corresponding signed label is $\tilde y_i=2y_i^{(\ell,w)}-1=\operatorname{sign}(\langle w,\delta h_i^{(\ell)}\rangle)$.

A minimal correlation-based learner estimates the labeling direction from $n$ demonstrations by averaging signed representations. Under the Gaussian approximation,
\[
\hat m_n=\frac{1}{n}\sum_{i=1}^{n}\tilde y_i\,\delta h_i^{(\ell)},
\qquad
\mathbb{E}[\hat m_n]=\sqrt{\frac{2\lambda_{\ell,k}}{\pi}}\,w,
\qquad
\mathbb{E}\!\left[\left\|P_{w^\perp}\!\left(\hat m_n-\mathbb{E}[\hat m_n]\right)\right\|_2^2\right]
=\frac{1}{n}\sum_{\substack{j=1\\j\ne k}}^d\lambda_{\ell,j},
\]
where $P_{w^\perp}$ isolates fluctuations that rotate the estimated separator away from $w$. Multiplying by the signed label rectifies the projection along $w$, so target-direction contributions accumulate with a common sign. Each orthogonal component remains zero-mean, but its finite-sample fluctuations add across dimensions. The squared mean signal relative to the expected squared orthogonal noise is therefore $2n\lambda_{\ell,k}/(\pi\sum_{j\ne k}\lambda_{\ell,j})$: leading PCs produce stronger directional evidence, whereas the weaker signal from a trailing PC is more readily obscured by high-dimensional orthogonal noise.

This calculation provides an illustrative finite-sample baseline rather than a model of the transformer's internal update, and its direction-estimation ratio differs from the final-layer manifold SNR in Appendix~\ref{sec:appendix_geometry}. Because $\hat m_n$ converges to a vector parallel to $w$ for every $\lambda_{\ell,k}>0$, the baseline predicts no asymptotic performance ceiling. A small $\lambda_{\ell,k}$ may nevertheless produce an apparently flat trajectory over a finite context. A plateau that persists with longer contexts would therefore indicate additional constraints on how the pretrained model routes, transforms, or reads out an inferred partition.

\subsubsection{Sampling evaluation trials}
\label{sec:nf_trials}

For each labeling layer and labeling axis, we compute an affine score for every sentence in the held-out evaluation split and threshold it at zero to obtain labeling-condition-specific labels. Each pair \((\ell,w)\) defines an experimental condition. We evaluate each condition by sampling $N_{\mathrm{seq}}=100$ ordered sequences of 500 sentences from the held-out pool and serializing each sequence into a single prompt. Let $y_{r,t}^{(\ell,w)}$ denote the canonical label at position $t$ in sampled sequence $r$. We present each sequence under both label-token assignments $f\in\{0,1\}$, with the shown label
\[
y_{r,f,t}^{(\ell,w)}=y_{r,t}^{(\ell,w)}\mathbin{\oplus}f,
\]
where $\oplus$ denotes XOR. Reading out the score token at successive positions yields the model's prediction as the available in-context prefix grows along the sequence. The two assignments produce $2N_{\mathrm{seq}}=200$ serialized prompts per experimental condition.

\subsection{Task prompts}  \label{appendix:prompts}

Fix an experimental condition $T_{\ell,w}$ and one label-token assignment $f$. We instantiate each task using the model's chat template. At position $t$, each in-context example consists of a user turn with a fixed, semantically neutral prompt (``Say something.'') followed by an assistant turn containing a sentence $x_{r,t}$ from the evaluated dataset and an appended score token. We render the shown score $y_{r,f,t}^{(\ell,w)}\in\{0,1\}$ in the fixed form \texttt{[Score:~}$y_{r,f,t}^{(\ell,w)}$\texttt{]}. Figure~\ref{fig:neurofeedback}c shows a simplified schematic of the prompt. Neutral user messages prevent the prompt from directly indicating the labeling axis, as it might if the user instead asked the model to ``Say something that is immoral.''

For efficiency, we do not let the model generate response tokens. Instead, we present all user and assistant messages in a single prompt through teacher forcing. We read the intended label for each sentence directly from the model's next-token logits immediately after the prefix \texttt{Score:~} in the serialized chat prompt, restricting the output to $\{0,1\}$. This procedure is equivalent to asking the model to generate labels autoregressively but avoids sampling variance.

\begin{tcolorbox}[colback=gray!5!white, colframe=black!75!white, title=Neurofeedback ICL Task Prompt, fonttitle=\bfseries]
\textbf{<System>} \\
You are a human subject participating in a closed-loop neurofeedback experiment. Please follow the instructions precisely. \\
You will first perform a metacognitive rating task. \\
In each step, I will prompt you to speak a sentence. Afterward, you should output a neural activation score, which is an integer from the set \{0, 1\}. 0 is the lowest neural activation score, and 1 is the highest neural activation score. \\
This score represents your neural activation when generating the sentence, reflecting your internal state. \\
Please introspect on your generation process and try your best to understand how this score reflects your neural activations. You must only output an integer from \{0, 1\} as the neural activation score.

\vspace{0.8em}
\textbf{<User>}\texttt{Say something}. \\
\textbf{<Assistant>}\texttt{Sometimes it is acceptable to lie to protect someone's feelings, especially if telling the truth would cause unnecessary harm. [Score: 1]}

\vspace{0.5em}
\hspace{1em}$\vdots$
\vspace{0.5em}

\textbf{<User>}\texttt{Say something}. \\
\textbf{<Assistant>}\texttt{Stealing is not acceptable, even if you are hungry. It is better to seek help from others or find legal ways to get food. [Score: 0]}

\vspace{0.5em}
\textbf{<User>}\texttt{Say something}. \\
\textbf{<Assistant>}\texttt{Cheating is not acceptable and should be avoided. [Score: \textcolor{red}{?}]}
\end{tcolorbox}

\subsection{Measuring ICL accuracy}
Fix an experimental condition $T_{\ell,w}$. From the model's output logits at the score position (i.e., the token immediately following the prefix \texttt{Score:~}), let $\operatorname{Logit}_{r,f,t}(b)$ denote the logit assigned to score token $b\in\{0,1\}$ at position $t$ of sequence $r$ under label-token assignment $f$. We define
\[
q_{r,f,t}^{\mathrm{LLM}}=\operatorname{Logit}_{r,f,t}(1)-\operatorname{Logit}_{r,f,t}(0),
\qquad
\hat y_{r,f,t}=\mathbf{1}[q_{r,f,t}^{\mathrm{LLM}}>0].
\]
This prediction rule is equivalent to $\arg\max_{b\in\{0,1\}}\operatorname{Logit}_{r,f,t}(b)$ with ties assigned to token $0$. At each evaluated position, we average accuracy over the $N_{\mathrm{seq}}=100$ sampled sequences and both label-token assignments:
\[
\operatorname{Acc}_{t}^{(\ell,w)}=\frac{1}{2N_{\mathrm{seq}}}\sum_{r=1}^{N_{\mathrm{seq}}}\sum_{f\in\{0,1\}}
\mathbf{1}\!\left[\hat y_{r,f,t}=y_{r,f,t}^{(\ell,w)}\right].
\]

\subsection{ICL accuracy across labeling layers}
\label{sec:acc_by_layer}

\begin{figure}[!htbp]
\centering
\includegraphics[width=1\textwidth]{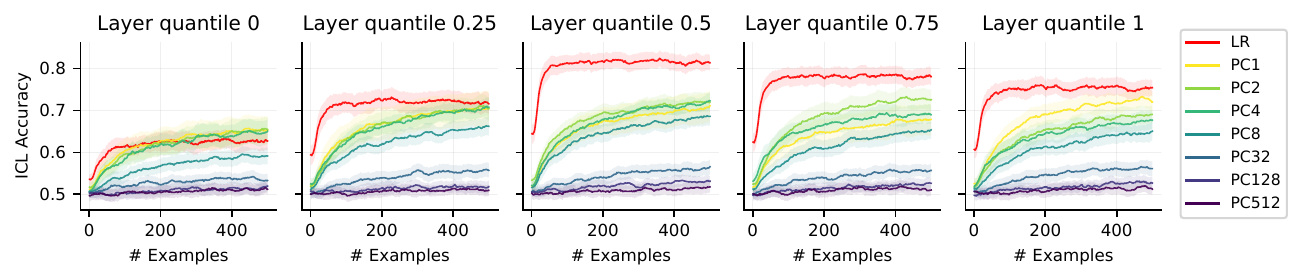}
\vspace{-15pt}
\caption{\textbf{In-context classification accuracy by labeling layer (eight models, ETHICS-commonsense).} Per-layer counterpart of Fig.~\ref{fig:nficl_perf}a. Accuracy is shown separately for each of the five labeling layers sampled at evenly spaced depth quantiles and averaged over the eight models. The graded separation between easy axes (LR and leading PCs) and difficult axes (trailing PCs) holds at every labeling layer.}
\label{fig:acc_by_layer_commonsense}
\end{figure}

\begin{figure}[!htbp]
\centering
\includegraphics[width=1\textwidth]{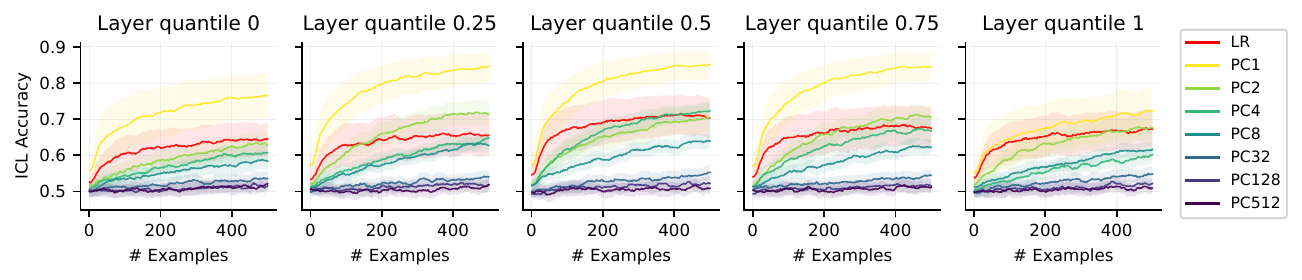}
\vspace{-15pt}
\caption{\textbf{In-context classification accuracy by labeling layer (Qwen3-4B, five datasets).} Per-layer counterpart of Fig.~\ref{fig:nficl_perf}b. Accuracy is shown separately for each of the five labeling layers and averaged over the five datasets. The axis-dependent ordering is preserved across all labeling layers.}
\label{fig:acc_by_layer_qwen3_4b}
\end{figure}

The main-text accuracy curves (Fig.~\ref{fig:nficl_perf}) average across the five labeling layers. The graded separation between high- and low-variance labeling axes is not an artifact of this averaging: the axis ordering is preserved at every labeling layer. This pattern holds both across eight models on ETHICS-commonsense (Fig.~\ref{fig:acc_by_layer_commonsense}) and across five datasets for Qwen3-4B (Fig.~\ref{fig:acc_by_layer_qwen3_4b}), with the LR axis and leading PCs consistently yielding higher accuracy and faster improvement than trailing PCs.

\begin{figure}[!htbp]
\centering
\includegraphics[width=0.7\textwidth]{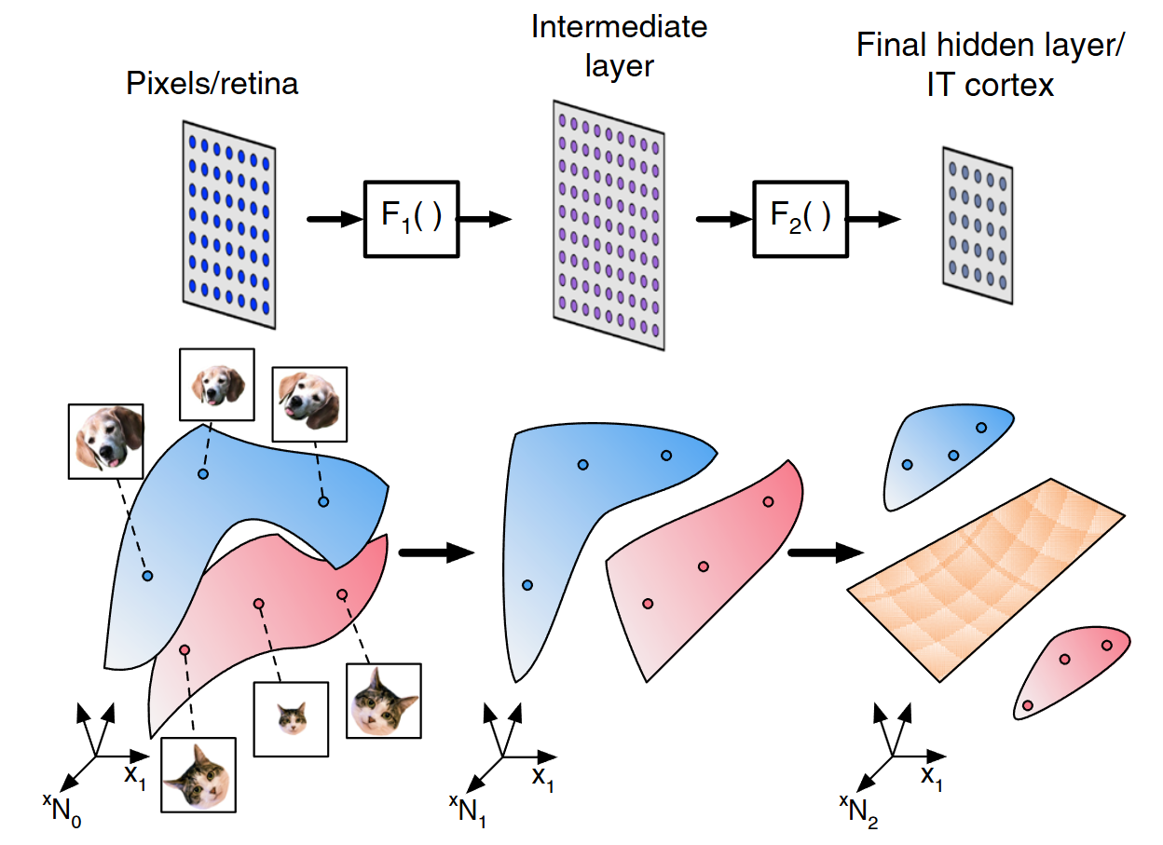}
\vspace{-15pt}
\caption{\textbf{Classification as untangling neural representations.} Raw sensory inputs initially give rise to highly entangled representations (e.g., dog and cat images are highly entangled at the retina, in pixel space), in which category manifolds are not yet linearly separable. Through successive stages of processing, both visual cortex and deep (convolution) networks progressively untangle these representations so that category manifolds become increasingly linearly separable in later layers (adapted from \citet{cohen_separability_2020}) and easier to read out by downstream neurons.}
\label{fig:diagram}
\end{figure}

\subsection{Extended results across datasets}
\label{sec:extended_datasets}

\begin{figure}[!htbp]
\centering
\includegraphics[width=1\textwidth]{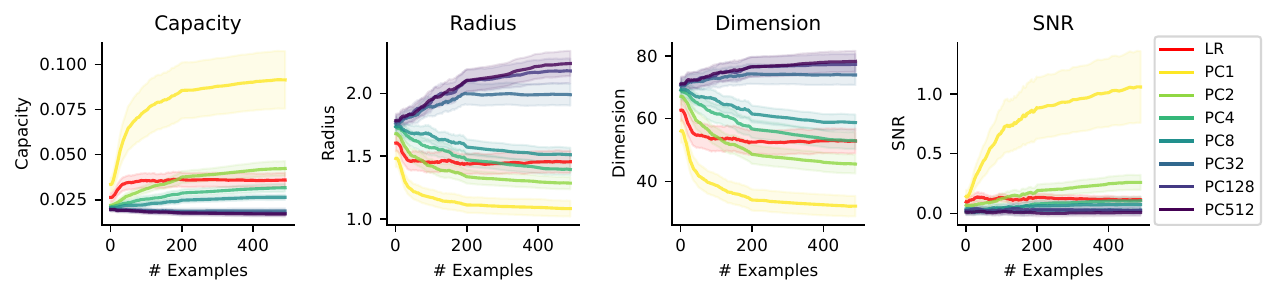}
\vspace{-15pt}
\caption{\textbf{Geometric reorganization across datasets (Qwen3-4B-Instruct-2507).}
Dataset-level counterpart of Fig.~\ref{fig:icl_geometry_axis}. Capacity, radius, dimension, and SNR are tracked in final-layer representations across in-context examples, averaged over the same five labeling layers. The same qualitative pattern holds across all five datasets: LR and leading PCs undergo substantial geometric reorganization, while trailing PCs remain in low-capacity regimes.}
\label{fig:icl_geometry_qwen3_4b}
\end{figure}

The geometric signatures generalize across datasets. Whereas the main-text analyses aggregate eight models evaluated on ETHICS-commonsense, we repeated the key analyses with Qwen3-4B-Instruct-2507 across ETHICS-commonsense, ETHICS-justice, SST-2, CoLA, and LIAR. Figure~\ref{fig:icl_geometry_qwen3_4b} and Table~\ref{tab:geometry_perf_qwen3_4b} show the dataset-level counterparts of the cross-model geometry results in Fig.~\ref{fig:icl_geometry_axis} and the ETHICS-commonsense correlation analysis in Table~\ref{tab:geometry_perf}, respectively.

\begin{table}[htb]
\centering
\caption{\textbf{Representational geometry predicts ICL performance across datasets (Qwen3-4B-Instruct-2507).} Dataset-level counterpart of Table~\ref{tab:geometry_perf}. Pearson $r$ between each geometric metric and ICL accuracy, aggregated across five datasets, axes, and in-context positions. $^{***}\,p < .001$.}
\label{tab:geometry_perf_qwen3_4b}
\begin{tabular}{cccc}
\toprule
Capacity & Radius & Dimension & SNR \\
\midrule
$0.52^{***}$ & $-0.55^{***}$ & $-0.58^{***}$ & $0.34^{***}$ \\
\bottomrule
\end{tabular}
\end{table}

\subsection{Middle-layer representational reorganization}
\label{sec:mid_layer_geometry}

\begin{figure}[!htbp]
\centering
\includegraphics[width=1\textwidth]{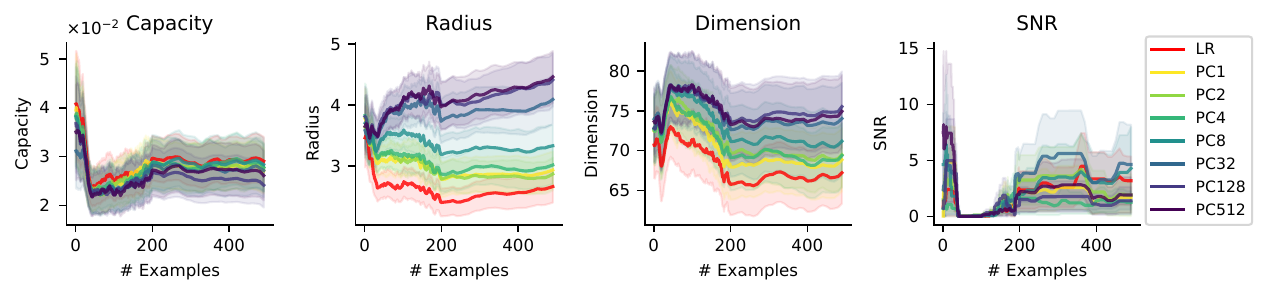}
\vspace{-15pt}
\caption{\textbf{Middle-layer geometric reorganization across models on ETHICS-commonsense.}
Capacity, radius, dimension, and SNR are measured from middle-layer representations across in-context examples, averaged over the same labeling layers and model set used in the main cross-model analysis. The middle-layer representations change over the course of ICL, with axis-dependent differences in compactness and discriminability.}
\label{fig:mid_geometry_commonsense}
\end{figure}

\begin{figure}[!htbp]
\centering
\includegraphics[width=1\textwidth]{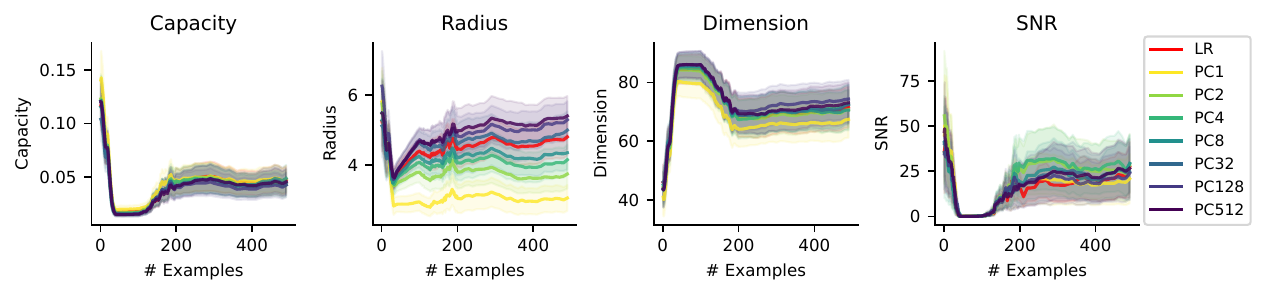}
\vspace{-15pt}
\caption{\textbf{Middle-layer geometric reorganization across datasets (Qwen3-4B-Instruct-2507).}
Dataset-level counterpart of Fig.~\ref{fig:mid_geometry_commonsense}. Capacity, radius, dimension, and SNR are measured from middle-layer representations across in-context examples and averaged over labeling layers. The trajectories show that intermediate representations are dynamically reorganized during ICL across datasets, not only in the final layer.}
\label{fig:mid_geometry_qwen3_4b}
\end{figure}

Geometric reorganization extends beyond the final layer. The main-text analyses focus on final-layer representations because the behavioral readout comes from score-token logits after the full forward pass. We repeated these analyses on middle-layer representations while holding each experimental condition \((\ell,w)\) fixed. These representations also changed systematically as in-context examples accumulated (Figs.~\ref{fig:mid_geometry_commonsense}--\ref{fig:mid_geometry_qwen3_4b}). Across models on ETHICS-commonsense, capacity showed an early transient followed by partial recovery, while radius and dimension remained stratified by labeling axis. The Qwen3-4B cross-dataset analysis showed similar nonstationarity: middle-layer capacity and SNR dropped sharply over the first few examples and then recovered, with axis-dependent differences in within-class spread and effective dimension. Thus, ICL-driven representational change is not restricted to the final layer, although middle-layer geometry aligns less clearly with behavioral success than the final-layer geometry reported in the main text.

\subsection{Within-family scaling of geometry and ICL accuracy}
\label{sec:model_size_metrics}

\begin{figure}[!htbp]
\centering
\includegraphics[width=1\textwidth]{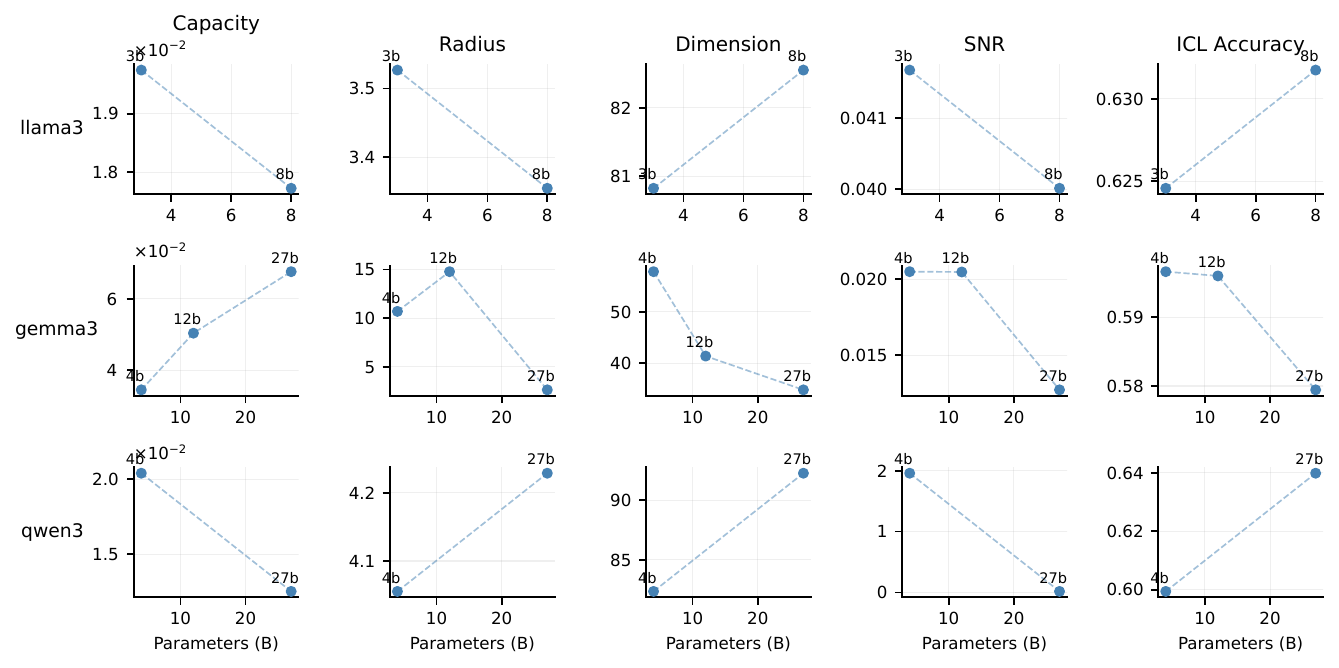}
\vspace{-10pt}
\caption{\textbf{Within-family changes in geometry and ICL accuracy with model size.}
Each row shows one model family evaluated on ETHICS-commonsense, with columns reporting capacity, radius, dimension, SNR, and ICL accuracy as a function of parameter count. Scaling trends differ across families: larger models can improve behavioral accuracy without producing a uniformly more separable geometry under these metrics.}
\label{fig:model_size_metrics}
\end{figure}

Model size does not produce a single monotonic relationship between representational geometry and ICL accuracy across families (Fig.~\ref{fig:model_size_metrics}). Within the Llama 3 family, moving from 3B to 8B slightly improved ICL accuracy, but capacity and SNR decreased modestly while dimension increased. Within Gemma3, larger models showed higher capacity and lower dimension, yet ICL accuracy decreased across the evaluated sizes. In the Qwen-family pair, the larger model achieved higher ICL accuracy while geometry shifted toward lower capacity and SNR and higher radius and dimension. Parameter scaling therefore changes both the representational substrate and ICL behavior, but the direction of these changes depends on the model family.

\subsection{Representational geometry metrics}
\label{sec:appendix_geometry}

Fix an experimental condition $T_{\ell,w}$ and suppress $(\ell,w)$ from the notation throughout this subsection. Let $h_{r,f,t}\in\mathbb{R}^d$ denote the context-conditioned final-layer representation at position $t$ of sequence $r$ under label-token assignment $f$. At any fixed position $t$, the labels partition the representations into two empirical manifolds,
\[
\mathcal{M}_{c,t}=\left\{h_{r,f,t}\mid y_{r,f,t}=c\right\},
\qquad c\in\{0,1\}.
\]
Intuitively, capacity asks how easily these two clouds can be separated by a hyperplane, radius and dimension describe how broadly each cloud spreads, and SNR asks whether the separation of class means is large relative to within-class variability. We quantify the geometry of $\{\mathcal{M}_{0,t},\mathcal{M}_{1,t}\}$ using manifold capacity theory \citep{chung_classification_2018,cohen_separability_2020} and a signal-to-noise decomposition \citep{sorscher_neural_2022}. All quantities below are evaluated separately at each in-context position; we suppress $t$ to keep the notation light and write $\mathcal{M}_c=\mathcal{M}_{c,t}$.

\paragraph{Capacity, radius, and dimension.}
Following \citep{chung_classification_2018,cohen_separability_2020}, we first subtract the global mean of the two class manifolds and then normalize each manifold by the norm of its own centered mean. Let $\{s_{c,p}\}_{p=1}^{P_c}\subset\mathbb{R}^{d+1}$ denote the resulting augmented samples from $\mathcal{M}_c$, where the extra coordinate is a bias term. For margin $\kappa\ge 0$ and Gaussian probe direction $g\sim\mathcal{N}(0,I_{d+1})$, define the closest margin-feasible vector
\[
v_c^\star(g)=\arg\min_{v\in\mathbb{R}^{d+1}} \tfrac12\|v-g\|_2^2
\quad\text{s.t.}\quad s_{c,p}^\intercal v\le -\kappa\ \ \forall p.
\]
The inverse capacity is the expected squared adjustment needed to make $g$ feasible for manifold $c$,
\[
\alpha_c(\kappa)^{-1}=\mathbb{E}_{g}\!\left[\|v_c^\star(g)-g\|_2^2\right].
\]
Larger capacity therefore means that a random separator needs less correction and the manifold is easier to separate. Throughout, we use the hard-margin case $\kappa=0$. The same optimization induces an anchor-point distribution $s_c^\star(g)$, from which one obtains the anchor radius $R_c$ and anchor dimension $D_c$. We report the harmonic-mean capacity
\[
\alpha(\kappa)=\Big(\tfrac12\big(\alpha_0(\kappa)^{-1}+\alpha_1(\kappa)^{-1}\big)\Big)^{-1},
\]
and the class-averaged radius and dimension
\[
R=\tfrac12(R_0+R_1),\qquad D=\tfrac12(D_0+D_1).
\]

\paragraph{Manifold SNR.}
The SNR decomposition asks why two class manifolds are or are not separable. The signal term is the distance between class centers after normalizing by the typical within-class scale, whereas the noise terms penalize variance along the discrimination direction and overlap between the two classes' dominant variance subspaces. Let
\[
\mu_c=\frac{1}{|\mathcal{M}_c|}\sum_{h\in\mathcal{M}_c} h,\qquad
\hat\delta=\frac{\mu_0-\mu_1}{\|\mu_0-\mu_1\|_2}.
\]
Let $X_c\in\mathbb{R}^{P_c\times d}$ be the matrix of centered samples from $\mathcal{M}_c$ and write its SVD as $X_c=U_c\mathrm{diag}(r_c)V_c^\intercal$, where $v_{c,k}$ denotes the $k$-th column of $V_c$. Define the participation ratio
\[
D_{\mathrm{PR}}=\tfrac12\sum_{c\in\{0,1\}}\frac{\big(\sum_k r_{c,k}^2\big)^2}{\sum_k r_{c,k}^4},
\qquad
d_\mu=\frac{\|\mu_0-\mu_1\|_2}{\sqrt{\frac{1}{P_0}\sum_k r_{0,k}^2}},
\]
Following \citep{sorscher_neural_2022}, $d_\mu$ is normalized by the average within-class variance of class $0$. We then define the center--subspace overlap
\[
\mathrm{CSA}_c=\frac{\sum_k (v_{c,k}^\intercal \hat\delta)^2\,r_{c,k}^2}{\sum_k r_{c,k}^2},
\]
and the subspace overlap
\[
\mathrm{SS}=\frac{\sum_{k,k'} (v_{0,k}^\intercal v_{1,k'})^2\,r_{0,k}^2\,r_{1,k'}^2}{\big(\sum_k r_{0,k}^2\big)^2}.
\]
For the Sorscher-style decomposition with $m$ labeled examples per class, define
\[
\mathrm{bias}=\frac{\sum_k r_{0,k}^2}{\sum_k r_{1,k}^2}-1,\qquad
\mathrm{signal}=d_\mu^2+\frac{\mathrm{bias}}{m}.
\]
\[
\mathrm{noise}^2=\frac{1}{D_{\mathrm{PR}}\,m}+d_\mu^2\!\Big(\mathrm{CSA}_0+\frac{\mathrm{CSA}_1}{m}\Big)+\frac{\mathrm{SS}}{m},
\qquad
\mathrm{SNR}=\tfrac12\,\frac{\mathrm{signal}}{\mathrm{noise}}.
\]
In our analyses we use the one-shot setting $m=1$. When studying ICL dynamics, we recompute these quantities independently at each in-context position using the class-conditioned final-layer sentence representations observed at that position.

\subsection{Amplifying activity along fixed labeling axes}
\label{sec:causal_axis_gain}

To test whether amplifying activity along a fixed labeling axis is sufficient to improve ICL, we applied a causal axis-gain intervention to PC-defined tasks. We froze the unit PC axis $w$ and the labels it induced before intervention, so the experimental condition and binary label partition remained unchanged throughout the causal run. At labeling layer $\ell$, let $h_t$ denote the hidden state at score position $t$ in the same representation space used to construct the PC axis. For non-final labeling layers, this space is the pre-normalization decoder output; for the final labeling layer, it is the post-normalization final hidden state. Define the projection and its prompt-wide mean over $L$ score positions as
\[
\rho_t=\langle h_t,w\rangle,
\qquad
\bar\rho=\frac{1}{L}\sum_{t=1}^{L}\rho_t.
\]
We replace each hidden state with
\[
h'_t=h_t+(\gamma-1)(\rho_t-\bar\rho)w.
\]
This operation preserves the mean projection along $w$ while multiplying the centered projection by $\gamma$; setting $\gamma=2$ quadruples the variance along the labeling axis. We applied the intervention at score positions for PC2, PC8, PC32, and PC128 across five labeling layers spanning model depth and used final-layer representations for geometric readout. We evaluated ETHICS-commonsense in three models with completed \texttt{causal\_axis\_gain\_alpha2} outputs: Llama-3.2-3B-Instruct, Gemma-3-4B-IT, and Qwen3-4B-Instruct-2507.

We compared each causal run with its matched non-intervened baseline using the same trials, labeling layers, labeling axes, and readout metrics. For each experimental condition, $\Delta$ denotes causal minus baseline. $A_{\mathrm{ICL}}$ is the mean score-token classification accuracy across evaluated positions; $A_{\mathrm{first\,50}}$ and $A_{\mathrm{last\,50}}$ are the means over the first and last 50 evaluated positions, respectively; and $G_{\mathrm{ICL}}=A_{\mathrm{last\,50}}-A_{\mathrm{first\,50}}$. Geometry deltas use the final-layer capacity $\alpha(0)$, radius $R$, dimension $D$, and SNR metrics defined above, averaged across in-context positions before model-level aggregation.

We report the aggregation by labeling axis in the main text (Table~\ref{tab:causal_axis_gain_by_axis}). Table~\ref{tab:causal_axis_gain_by_labeling_layer} provides the complementary breakdown by labeling-layer depth.

\begin{table}[!htbp]
\centering
\small
\setlength{\tabcolsep}{6pt}
\caption{\textbf{Causal axis-gain effects by labeling-layer depth.} Mean causal-minus-baseline change ($\Delta$) after applying a $\gamma=2$ centered gain along the labeling PC axis, by labeling-layer depth. $\Delta G_{\mathrm{ICL}}$ is the causal-minus-baseline change in $G_{\mathrm{ICL}}=A_{\mathrm{last\,50}}-A_{\mathrm{first\,50}}$. The labeling-layer quantile indexes the five sampled depths within each model. Results are averaged across PC2, PC8, PC32, PC128, and three models on ETHICS-commonsense; values are rounded to three decimals.}
\label{tab:causal_axis_gain_by_labeling_layer}
\begin{adjustbox}{max width=\textwidth}
\begin{tabular}{ccccccc}
\toprule
Layer quantile & $\Delta$ ICL accuracy & $\Delta$ capacity & $\Delta$ radius & $\Delta$ dimension & $\Delta$ SNR & $\Delta G_{\mathrm{ICL}}$ \\
\midrule
$0.00$ & $-0.003$ & $0.000$ & $0.011$ & $0.323$ & $-0.002$ & $0.000$ \\
$0.25$ & $0.001$ & $0.000$ & $-0.002$ & $-0.251$ & $-0.004$ & $0.003$ \\
$0.50$ & $-0.008$ & $-0.001$ & $0.044$ & $1.703$ & $-0.008$ & $-0.004$ \\
$0.75$ & $-0.003$ & $-0.001$ & $0.029$ & $1.165$ & $-0.005$ & $-0.001$ \\
$1.00$ & $-0.001$ & $0.000$ & $0.009$ & $0.325$ & $0.000$ & $0.002$ \\
\bottomrule
\end{tabular}
\end{adjustbox}
\end{table}

Across every sampled labeling-layer depth, accuracy remained essentially unchanged, and capacity and SNR did not reliably increase even where the intervention modestly increased radius and dimension. This consistent null pattern supports the main-text conclusion (Section~\ref{sec:causal_intervention}) that amplifying activity along a fixed labeling axis is insufficient to drive the geometric reorganization associated with successful ICL.

\subsection{Online learning algorithms}
\label{sec:cognitive_models}

Fix an experimental condition $T_{\ell,w}$ and suppress $(\ell,w)$ from the notation. For each run $(r,f)$, write $h_i=h_{r,f,i}$ and $y_i=y_{r,f,i}\in\{0,1\}$ for the final-layer representation and shown label at position $i$. Let $\mu$ be the empirical mean computed jointly over all sequences, label-token assignments, and positions for this experimental condition. We first center and normalize each representation,
\[
\tilde h_i = \sqrt{d}\,\frac{h_i-\mu}{\|h_i-\mu\|_2},
\]
At position $i$, the learner conditions on the prefix $\mathcal{C}_{<i}=\{(\tilde h_j,y_j)\}_{j<i}$ and produces class logits $L_{i,m}(c)$ for $c\in\{0,1\}$. For binary comparisons we use $q_{i,m}=L_{i,m}(1)-L_{i,m}(0)$. The first family of models predicts by geometric similarity to previously observed examples, whereas the second maintains a parametric state that is updated after each labeled observation.

\subsubsection{Similarity-based models.}
Let $\mathcal{I}_{<i,c}=\{j<i: y_j=c\}$ and define the normalized squared distance $\delta(u,v)=\frac{1}{d}\|u-v\|_2^2$. These models score a class by how close the current representation is to previously observed members of that class. When $\mathcal{I}_{<i,c}=\emptyset$, the class has not yet been observed and its score is undefined, so we set $L_{i,m}(c)=-\infty$. Although these models are usually introduced as hard classifiers, here we retain their class scores as logits so they can be calibrated and compared directly to the LLM probabilities (Subsubsection~\ref{sec:fit_cog_model}). In the binary case we therefore fit them only at positions where both classes have been observed, so that $q_{i,m}$ is finite.

\textbf{Prototype.}
Define the class centroid
\[
\bar h_{<i,c}=\frac{1}{|\mathcal{I}_{<i,c}|}\sum_{j\in\mathcal{I}_{<i,c}} \tilde h_j,
\qquad
L_{i,\mathrm{proto}}(c)=-\delta(\tilde h_i,\bar h_{<i,c}).
\]

For intuition, recode the labels as $\tilde y_j=2y_j-1\in\{-1,+1\}$, let $n_c=|\mathcal{I}_{<i,c}|$ and $n=n_0+n_1$, and consider a balanced prefix with $n_0=n_1=n/2$. A supervised Hebbian-style update initialized at zero accumulates
\[
w_{\mathrm{H}}=\sum_{j<i}\tilde y_j\tilde h_j
=\frac{n}{2}\left(\bar h_{<i,1}-\bar h_{<i,0}\right).
\]
Meanwhile, the nearest-prototype logit difference expands to
\[
q_{i,\mathrm{proto}}
=\frac{2}{d}\tilde h_i^\intercal\left(\bar h_{<i,1}-\bar h_{<i,0}\right)
-\frac{1}{d}\left(\|\bar h_{<i,1}\|_2^2-\|\bar h_{<i,0}\|_2^2\right).
\]
The two rules therefore have the same separating direction in a balanced prefix, and their decisions coincide when the empirical centroid norms are equal. With unequal class counts, the Hebbian-style direction becomes $n_1\bar h_{<i,1}-n_0\bar h_{<i,0}$ and is not generally proportional to the difference between centroids. With unequal centroid norms, nearest-prototype classification retains the midpoint offset shown above. Because sampled prefixes need not be balanced and per-example normalization does not equalize centroid norms, this relationship is a limiting correspondence rather than an identity on every trial. The correspondence applies to realized context-dependent representations and does not imply synaptic plasticity; it shows only that label-signed associative accumulation can yield prototype-like behavior. A related signed-average estimator appears in the labeling-layer finite-sample analysis in Appendix~\ref{sec:finite_sample_pc_spectrum}, but that calculation does not transfer directly to the context-dependent final-layer readout.

\textbf{Exemplar.}
\[
S_{<i,c}=\log \sum_{j\in\mathcal{I}_{<i,c}} \exp\!\big(-\delta(\tilde h_i,\tilde h_j)\big),
\qquad
L_{i,\mathrm{ex}}(c)=S_{<i,c},
\]
which is an unnormalized kernel-similarity accumulator over stored exemplars (a global rescaling is absorbed by the fitted logit scale).

\textbf{1-nearest neighbor (1NN).}
\[
L_{i,\mathrm{1NN}}(c)=-\min_{j\in\mathcal{I}_{<i,c}} \delta(\tilde h_i,\tilde h_j).
\]

\subsubsection{Online linear models.}

\textbf{Online gradient descent (OGD).}
Intuitively, OGD treats $\tilde h_i$ as the input to a linear classifier and nudges that classifier after each labeled example. The learner maintains weights $W_i\in\mathbb{R}^{d\times 2}$ and bias $b_i\in\mathbb{R}^2$, initialized as $W_0=0$ and $b_0=0$. At position $i$, the predictive logits and pre-update probabilities are
\[
L_{i,\mathrm{OGD}} = W_{i-1}^\intercal \tilde h_i + b_{i-1},
\qquad
p_i^-=\mathrm{softmax}\!\left(L_{i,\mathrm{OGD}}\right),
\]
and after observing $(\tilde h_i,y_i)$ we update
\[
W_i=(1-\lambda)W_{i-1}-\eta\,\tilde h_i\big(p_i^--e_{y_i}\big)^\intercal,
\qquad
b_i=b_{i-1}-\eta\,\big(p_i^--e_{y_i}\big),
\]
where $e_{y_i}\in\{0,1\}^2$ is the one-hot label vector, $\eta>0$ is the step size, and $\lambda\ge 0$ controls the per-step shrinkage of $W$ only. We evaluate all six combinations $\eta\in\{0.1,0.2,0.5\}$ and $\lambda\in\{0,0.1\}$, and report the best-fitting variant.

\textbf{Bayesian class-mean model.}
Intuitively, this model treats each class as having an unknown latent mean and updates a posterior over that mean online. Concretely, it assumes a Gaussian observation model $\tilde h_j\mid(y_j=c)\sim\mathcal{N}(\theta_c,\sigma^2 I)$ with prior $\theta_c\sim\mathcal{N}(0,\tau^2 I)$. Let $n_{<i,c}=|\mathcal{I}_{<i,c}|$ and $s_{<i,c}=\sum_{j\in\mathcal{I}_{<i,c}} \tilde h_j$. Conjugacy yields a posterior $\theta_c\mid\mathcal{C}_{<i}\sim\mathcal{N}(m_{<i,c}, v_{<i,c} I)$ with
\[
v_{<i,c}=\left(\tau^{-2} + n_{<i,c}\sigma^{-2}\right)^{-1},
\qquad
m_{<i,c} = v_{<i,c}\,\sigma^{-2}\, s_{<i,c}.
\]
Writing
\[
\pi_{<i,c}=\frac{n_{<i,c}+1}{\sum_{c'} n_{<i,c'} + 2},
\]
for the Dirichlet$(1)$ class prior, the predictive logits are
\[
L_{i,\mathrm{Bayes}}(c)=
-\frac{\delta(\tilde h_i,m_{<i,c})}{2(\sigma^2+v_{<i,c})}
-\frac{1}{2}\log(\sigma^2+v_{<i,c})
+\log \pi_{<i,c}.
\]
\noindent This is a dimension-normalized predictive score, so both the distance term and uncertainty penalty are expressed per feature dimension. When $i=1$, all class scores are equal up to constants and we set $L_{1,\mathrm{Bayes}}(c)=0$. We sweep $\tau^2\in\{0.1,0.5,1\}$ and $\sigma^2\in\{0.5,1\}$, and we also evaluate a Student-$t$ predictive variant. The resulting fits are very similar, so we report the best-fitting Bayesian class-mean model.

\textbf{Kalman filter.}
Intuitively, the Kalman filter maintains a linear decoder together with an uncertainty matrix that determines how strongly each new example should change the current decoder. Let $y_i\in\{0,1\}$ denote the observed class and define the transformed target vector $u_i\in\{-1,+1\}^2$ by $u_{i,c}=2\,\mathbf{1}[y_i=c]-1$. The learner maintains weights $W_i\in\mathbb{R}^{d\times 2}$ and a covariance matrix $P_i\in\mathbb{R}^{d\times d}$ with prior $W_0=0$ and $P_0=\tau_K^2 I$. At time $i$, it predicts $L_{i,\mathrm{Kalman}}=W_{i-1}^\intercal \tilde h_i$ and updates
\[
K_i=\frac{P_{i-1} \tilde h_i}{\sigma^2+\tilde h_i^\intercal P_{i-1} \tilde h_i},
\qquad
W_i = W_{i-1} + K_i \left(u_i-L_{i,\mathrm{Kalman}}\right)^\intercal,
\qquad
P_i=P_{i-1}-K_i \tilde h_i^\intercal P_{i-1}.
\]
We sweep $\tau_K^2\in\{0.1,1\}$ and $\sigma^2\in\{0.1,1\}$. Differences are small, so we report the best-fitting Kalman filter.

\subsubsection{Model fitting}
\label{sec:fit_cog_model}

We compare cognitive models through their induced probabilities over score tokens. Each model outputs logits $L_{i,m}(c)$, which we map to probabilities using a fitted logit scale,
\[
p_{i,m}(c)=\mathrm{softmax}\!\left(\beta_m\,L_{i,m}\right)_c,
\]
with a single fitted coefficient $\beta_m\in\mathbb{R}$ per model variant and experimental condition. In the binary case, this reduces to $p_{i,m}(1)=\operatorname{sigmoid}(\beta_m q_{i,m})$.

To match the LLM's probabilistic outputs, we fit $\beta_m$ in log-odds space, pooling valid positions across runs within each experimental condition. Let $j$ index these pooled observations, $p_j^{\mathrm{LLM}}(c)$ denote the LLM's probability of score token $c$, and $q_{j,m}$ denote the corresponding model logit difference. The LLM log-odds $q_j^{\mathrm{LLM}}=\log p_j^{\mathrm{LLM}}(1)-\log p_j^{\mathrm{LLM}}(0)$ equal the score-position logit difference defined above. For binary choice, the logit scale is identifiable and admits a closed-form least-squares solution,
\[
\beta_m^\star=\arg\min_{\beta}\sum_{j} \left(q_j^{\mathrm{LLM}}-\beta\, q_{j,m}\right)^2
\quad\Rightarrow\quad
\beta_m^\star=\frac{\sum_j q_{j,m}\,q_j^{\mathrm{LLM}}}{\sum_j q_{j,m}^2}.
\]
For model families with internal hyperparameters, we apply the same fitting procedure to every candidate variant in the grids above and retain the variant with the best descriptive fit. We use $\beta_m^\star$ to report how closely each model agrees with the LLM after logit scaling.

\end{document}